# Preparing Korean Data for the Shared Task on Parsing Morphologically Rich Languages

Jinho D. Choi
Department of Computer Science
University of Massachusetts Amherst
jdchoi@cs.umass.edu

## Abstract

This document gives a brief description of Korean data prepared for the SPMRL 2013 shared task (Seddah et al., 2013). A total of 27,363 sentences with 350,090 tokens are used for the shared task. All constituent trees are collected from the KAIST Treebank and transformed to the Penn Treebank style. All dependency trees are converted from the transformed constituent trees using heuristics and labeling rules designed specifically for the KAIST Treebank. In addition to the gold-standard morphological analysis provided by the KAIST Treebank, two sets of automatic morphological analysis are provided for the shared task, one is generated by the HanNanum morphological analyzer, and the other is generated by the Sejong morphological analyzer.

## 1 Constituent Treebank

All constituent trees are collected from the KAIST Treebank (Choi et al., 1994). The KAIST Treebank contains about 31K manually annotated constituent trees from 97 different sources (e.g., newspapers, novels, textbooks). After filtering out trees with annotation errors, a total of 27,363 trees with 350,090 tokens are collected. Table 1 shows distributions of the training, development, and evaluation sets used for the shared task.

|        | Train   | Develop | Evaluate |
|--------|---------|---------|----------|
| Trees  | 23,010  | 2,066   | 2,287    |
| Tokens | 296,446 | 25,278  | 28,366   |

Table 1: Distributions of the training, development, and evaluation sets used for the shared task.

Constituent trees in the KAIST Treebank also come with manually inspected morphological analysis based on 'eojeol'. An eojeol contains root-forms of word tokens agglutinated with grammatical affixes (e.g., case particles, ending markers). An eojeol can consist of more than one word token; for instance, a compound noun "*bus stop*" is often represented as one eojeol in Korean, 버스정류장, which can be broken into two word tokens, 버스 (bus) and 정류장 (stop). Each eojeol in the KAIST Treebank is separated by white spaces regardless of punctuation. Figure 1 shows morphological analysis for a sentence, "*I drank cognac.*" in Korean.

| 나는 | 꼬냑(Cognac)을 | 들이켰다. |
|------|----------------|-----------|
| I    | cognac         | drank     |

⇩ ⇩ ⇩

| 나+는  | 꼬냑+(+Cognac+)+을        | 들이키+었+다+. |
|--------|---------------------------|-----------------|
| I+*tpc* | cognac+(+Cognac+)+*obj* | drink+*past*+*final*+. |

Figure 1: Morphological analysis for a sentence, "*I drank cognac.*" in Korean, where each morpheme is separated by a plus sign (+). *tpc*: topical auxiliary, *obj*: objective case particle, *past*: past-tense ending marker, *final*: final ending marker.

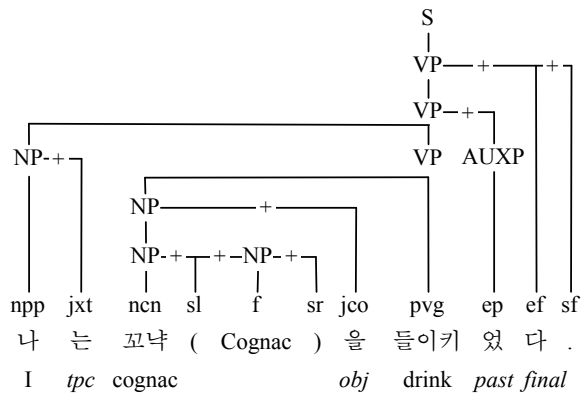

Figure 2: A constituent tree provided by the KAIST Treebank for the sentence in Figure 1.

There are 3 eojeols in this sentence. The 2nd eojeol, "꼬냑(Cognac)을", consists of a Korean word for cognac, 꼬냑, a parenthetical notation, (Cognac), and an object case particle, 을. Each morpheme is separated by a plus sign (+). Figure 2 shows a constituent tree provided by the KAIST Treebank for the same sentence. Some morphemes are not children but agglutinated to their parent phrases. For instance, 나 (*I*) is a child of an `NP`, but the topical auxiliary, 는 (*tpc*), is not a child but agglutinated to the `NP` in the KAIST Treebank (these relations are represented with plus signs in Figure 2; see Lee et al. (1997) for more details about the KAIST Treebank bracketing guidelines).

| | | | | | |
|---|---|---|---|---|---|
| `nbn` | Non-unit bound noun | `mad` | Demonstrative adverb | `jxf` | Final auxiliary |
| `nbu` | Unit bound noun | `mag` | General adverb | `jxt` | Topical auxiliary |
| `ncn` | Non-predicative common noun | `maj` | Conjunctive adverb | `ecc` | Coordinate conjunction EM |
| `ncpa` | Active-predicative common noun | `mma` | Attributive adnoun | `ecs` | Subordinate conjunction EM |
| `ncps` | Stative-predicative common noun | `mmd` | Demonstrative adnoun | `ecx` | Auxiliary conjunction EM |
| `nnc` | Cardinal numerals | `jca` | Adverbial CP | `ef` | Final EM |
| `nno` | Ordinal numerals | `jcc` | Complemental CP | `ep` | Pre-final EM |
| `npd` | Demonstrative pronoun | `jcj` | Conjunctive CP | `etm` | Adnominalizing EM |
| `npp` | Personal pronoun | `jcm` | Adnominal CP | `etn` | Nominalizing EM |
| `nq` | Proper noun | `jco` | Objective CP | `xp` | Prefix |
| `f` | Foreign word | `jcr` | Quotative CP | `xsa` | Adverb DS |
| `paa` | Attributive adjective | `jcs` | Subjective CP | `xsm` | Adjective DS |
| `pad` | Demonstrative adjective | `jct` | Comitative CP | `xsn` | Noun DS |
| `pvd` | Demonstrative verb | `jcv` | Vocative CP | `xsv` | Verb DS |
| `pvg` | General verb | `jp` | Predicative maker | `ii` | Interjection |
| `px` | auxiliary verb | `jxc` | common auxiliary | `sd, sf, sl, sp, sr, su, sy` | |

Table 2: POS tags in the KAIST Treebank (CP: case particle, EM: ending marker, DS: derivational suffix, `sd, sf, sl, sp, sr, su, sy`: different types of punctuation).

Tables 2 and 3 show part-of-speech tags and phrase types in the KAIST Treebank (see Kim et al. (2003) for more details about these tags).

| Type | Description |
|---|---|
| `ADJP` | Adjective phrase |
| `ADVP` | Adverb phrase |
| `AUXP` | Auxiliary verb phrase |
| `IP` | Interjection phrase |
| `NP` | Noun phrase |
| `VP` | Verb phrase |
| `S` | Sentence |

Table 3: Phrase types in the KAIST Treebank.

## 2 Dependency Treebank

All dependency trees are automatically converted from the constituent trees in Section 1. Unlike English that requires complicated head-finding rules to find the head of each phrase (Choi and Palmer, 2012), Korean is a head final language such that the rightmost constitutent in each phrase becomes the head of the phrase. To make our dependency structure more semantically oriented, the following cases are not considered heads unless they are the only constituents in phrases.

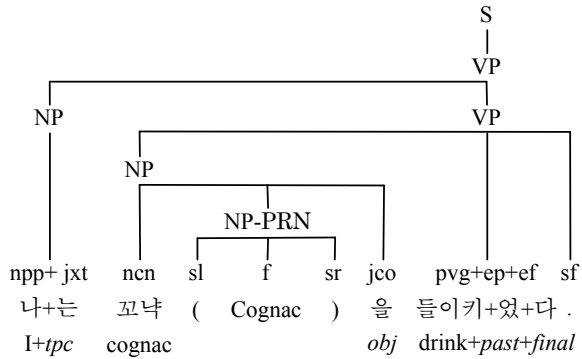

Figure 3: A constituent tree provided by the KAIST Treebank for the sentence in Figure 1.

Figure 3 shows a Penn Treebank style constituent tree transformed from the tree in Figure 2. Following the Penn Korean Treebank guidelines (Han et al., 2002), punctuation is separated as individual tokens and parenthetical notations surrounded by round brackets are grouped into individual phrases with a function tag, `PRN`. This function tag is useful for the dependency conversion in Section 2 because the parenthetical notation, (Cognac), should be a dependent of the head word, 꼬냑, regardless of its part-of-speech tag or position in the phrase.

1. A constituent is `AUXP` or `IP`.

2. A constituent has a function tag `PRN`.

3. A constituent consists of only gramatical affixes (`j*`, `e*`, or `x*` in Table 2).

4. A constituent consists of only punctuation (`s*` in Table 2).

Given these heuristics, the constituent tree in Figure 3 can be converted into the dependency tree in Figure 4. The root of this tree is *drink+past+final*, which takes *I+tpc* and *cognac+(...)+obj* as its subject and object, respectively.

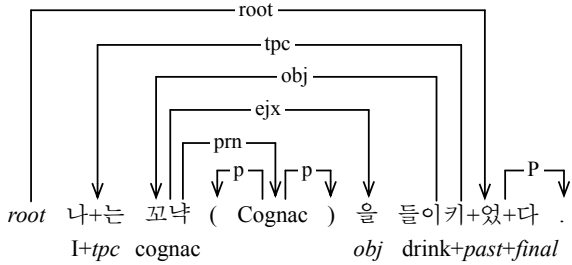

Figure 4: A dependency tree converted from the constituent tree in Figure 3. See Table 4 for details about dependency labels.

Unlike the Stanford dependency where the leftmost conjunct becomes the head of all other conjuncts and conjunctions in a coordination phrase (de Marneffe and Manning, 2008), the rightmost conjunct becomes the head in our dependency structure, which aligns well with our head final analogy. In Figure 5, *bacteria+obj* becomes the heads of a conjunct *cell* and a conjunction *and*, and *kill+final* becomes the head of a conjunct *destroy+cc*.

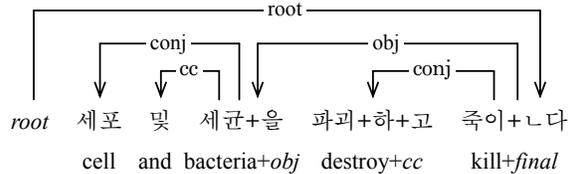

Figure 5: A dependency tree containing coordination. *obj*: objective case particle, *cc*: coordinating conjunction, *final*: final ending marker.

Constituent trees in the KAIST Treebank do not consist of function tags indicating syntactic or semantic roles, which makes it difficult to generate dependency labels. However, it is possible to generate meaningful labels by using rich morphology in Korean. For instance, case particles in Table 2 give good indications of what syntactic roles eojeols with such particles should take. Given this analogy, dependency labels are generated as in Table 4.

## 3 Automatic morphological analysis

Two sets of automatic morphological analysis are provided for the shared task. One is generated by the HanNanum morphological analyzer, which is developed by the KAIST Semantic Web Research Center.[1] The HanNanum morphological analyzer gives the same morphemes and POS tags as in Table 2. The other is generated by the Sejong morphological analyzer (지능형 형태소 분석기), which is developed by the Sejong Project.[2] The Sejong morphological analyzer gives a different set of morphemes and POS tags as described in Table 5.

Besides fine-grained POS tags in Tables 2 and 5, coarse-grained POS tags are also provided for both constituent and dependency trees (the `cpos` field in the Penn Treebank format and the 4th column in the CoNLL format). The coarse-grained POS tags are mostly represented by the first characters of the fine-grained POS tags except for SN, NF, SL, SH, NV, which are represented as N, N, F, F, and V.

| Case labels | |
|---|---|
| `comit` | Comitative |
| `comp` | Complement |
| `obj` | Object |
| `quot` | Quotation |
| `sbj` | Subject |
| `tpc` | Topic |
| **Inferred labels** | |
| `adn` | Adnominal modifier |
| `adv` | Adverbial modifier |
| `amod` | Modifier of adjective |
| `aux` | Auxiliary verb |
| `cc` | Coordinating conjunction |
| `conj` | Conjunct |
| `dep` | Unclassified dependent |
| `ejx` | Gramatical affixes |
| `intj` | Interjection |
| `nmod` | Modifier of nominal |
| `p` | Punctuation |
| `prn` | Parenthetical notation |
| `root` | Root |
| `sub` | Subordination |
| `vmod` | Modifier of predicate |

Table 4: Dependency labels.

## Acknowledgments

Special thanks to Prof. Key-Sun Choi at KAIST and Dr. Jungyeul Park at an Amzer Vak for kindly providing the KAIST Treebank.

---

[1] http://kldp.net/projects/hannanum
[2] http://www.sejong.or.kr

| | | | | | | | |
|---|---|---|---|---|---|---|---|
| `NNG` | General noun | `MM` | Adnoun | `EP` | Prefinal EM | `JX` | Auxiliary PR |
| `NNP` | Proper noun | `MAG` | General adverb | `EF` | Final EM | `JC` | Conjunctive PR |
| `NNB` | Bound noun | `MAJ` | Conjunctive adverb | `EC` | Conjunctive EM | `IC` | Interjection |
| `NP` | Pronoun | `JKS` | Subjective CP | `ETN` | Nominalizing EM | `SN` | Number |
| `NR` | Numeral | `JKC` | Complemental CP | `ETM` | Adnominalizing EM | `SL` | Foreign word |
| `VV` | Verb | `JKG` | Adnomial CP | `XPN` | Noun prefix | `SH` | Chinese word |
| `VA` | Adjective | `JKO` | Objective CP | `XSN` | Noun DS | `NF` | Noun-like word |
| `VX` | Auxiliary predicate | `JKB` | Adverbial CP | `XSV` | Verb DS | `NV` | Predicate-like word |
| `VCP` | Copula | `JKV` | Vocative CP | `XSA` | Adjective DS | `NA` | Unknown word |
| `VCN` | Negation adjective | `JKQ` | Quotative CP | `XR` | Base morpheme | `SF, SP, SS, SE, SO, SW` | |

Table 5: POS tags generated by the Sejong morphological analyzer (CP: case particle, EM: ending marker, DS: derivational suffix, PR: particle, `SF SP SS SE SO`: different types of punctuation).

## References

Jinho D. Choi and Martha Palmer. 2012. Guidelines for the Clear Style Constituent to Dependency Conversion. Technical Report 01-12, University of Colorado Boulder.

Key-sun Choi, Young S. Han, Young G. Han, and Oh W. Kwon. 1994. KAIST Tree Bank Project for Korean: Present and Future Development. In *In Proceedings of the International Workshop on Sharable Natural Language Resources*, pages 7–14.

Marie-Catherine de Marneffe and Christopher D. Manning. 2008. The Stanford typed dependencies representation. In *Proceedings of the COLING workshop on Cross-Framework and Cross-Domain Parser Evaluation*.

Chung-hye Han, Na-Rae Han, Eon-Suk Ko, Martha Palmer, and Heejong Yi. 2002. Penn korean treebank: Development and evaluation. In *In Proceedings of the 16th Pacific Asia Conference on Language, Information and Computation*, PACLIC'02.

Seongyong Kim, Key-sun Choi, and Kong Joo Lee. 2003. Automatic Generation of Composite Labels Using Part-of-Speech Tags for Parsing Korean. *International Journal of Computer Processing of Languages*, 16(3):197–218.

Kong Joo Lee, Byung-Gyu Chang, and Gil Chang Kim. 1997. Bracketing Guidelines for Korean Syntactic Tree Tagged Corpus. Technical Report CS/TR-97-112, Department of Computer Science, KAIST.

Djamé Seddah, Reut Tsarfaty, Sandra Kübler, Marie Candito, Jinho D. Choi, Richárd Farkas, Jennifer Foster, Iakes Goenaga, Koldo Gojenola, Yoav Goldberg, Spence Green, Nizar Habash, Marco Kuhlmann, Wolfgang Maier, Joakim Nivre, Adam Przepiorkowski, Ryan Roth, Wolfgang Seeker, Yannick Versley, Veronika Vincze, Marcin Woliński, Alina Wróblewska, and Eric Villemonte de la Clérgerie. 2013. Overview of the SPMRL 2013 Shared Task: A Cross-Framework Evaluation of Parsing Morphologically Rich Languages. In *Proceedings of the 4th Workshop on Statistical Parsing of Morphologically Rich Languages: Shared Task*, SPMRL'13.